\documentclass[sigconf]{acmart}
\AtBeginDocument{%
  }

\setcopyright{acmlicensed}
\copyrightyear{2025}

\acmYear{2025}
\acmConference[ACM Multimedia]{}{October 27--31, 2025}{Dublin, Ireland}
\acmISBN{978-1-4503-XXXX-X/2018/06}

\begin{document}

\title{Behave Your Motion: Habit-preserved Cross-category Animal Motion Transfer}

%
% The "author" command and its associated commands are used to define
% the authors and their affiliations.
% Of note is the shared affiliation of the first two authors, and the
% "authornote" and "authornotemark" commands
% used to denote shared contribution to the research.
\author{Zhimin Zhang}
\authornote{Both authors contributed equally to this research.}
% \email{zm_zhang@stu.pku.edu.cn}
\author{Bi'an Du}
\authornotemark[1]
% \email{pkudba@stu.pku.edu.cn}
\affiliation{%
  \institution{Wangxuan Institute of Computer Technology, Peking University}
  \city{Beijing}
  \country{China}
}

\author{Caoyuan Ma}
\author{Zheng Wang}
\affiliation{%
  \institution{Wuhan University}
  \city{Wuhan}
  \country{China}}

\author{Wei Hu}
\authornote{Corresponding author.}
\affiliation{%
  \institution{Wangxuan Institute of Computer Technology, Peking University}
  \city{Beijing}
  \country{China}
}

\renewcommand{\shortauthors}{Trovato et al.}

\begin{abstract}
Animal motion embodies species-specific behavioral habits, making the transfer of motion across categories a critical yet complex task for applications in animation and virtual reality. 
Existing motion transfer methods, primarily focused on human motion, emphasize skeletal alignment (motion retargeting) or stylistic consistency (motion style transfer), often neglecting the preservation of distinct habitual behaviors in animals.
To bridge this gap, we propose a novel habit-preserved motion transfer framework for cross-category animal motion. 
Built upon a generative framework, our model introduces a habit-preservation module with category-specific habit encoder, allowing it to learn motion priors that capture distinctive habitual characteristics. 
Furthermore, we integrate a large language model (LLM) to facilitate the motion transfer to previously unobserved species.
To evaluate the effectiveness of our approach, we introduce the DeformingThings4D-skl dataset, a quadruped dataset with skeletal bindings, and conduct extensive experiments and quantitative analyses, which validate the superiority of our proposed model.

\end{abstract}    

\thanks{This paper is supported by the State Key Laboratory of General Artificial Intelligence.}





\begin{teaserfigure}
    \centering
    \vspace{-20pt}
    \includegraphics[width=1\linewidth]{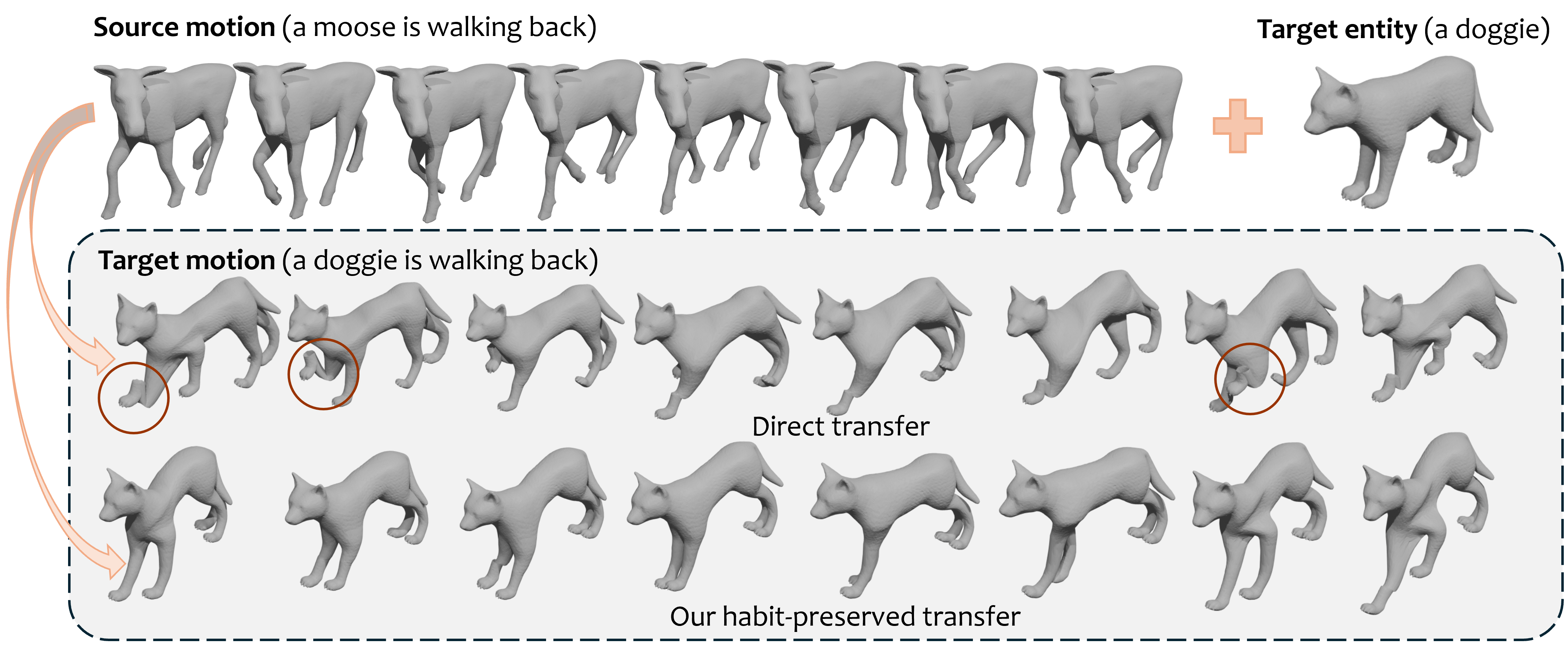}
    \captionof{figure}{\textbf{Visual results of motion transfer methods.} The same "walking back" motion should appear differently across species: for example, moose with their long limbs and doggies with their short, compact bodies have distinct muscle and skeletal structures, resulting in different bending angles in their movements. Directly transferring motion of a moose to a doggie can disrupt natural movement patterns. In cross-category motion transfer, it is crucial to account for each species' unique habits.}
    \label{fig:teaser}
\end{teaserfigure}%


\maketitle

% \let\thefootnote\relax\footnotetext{This paper is supported by the State Key Laboratory of General Artificial Intelligence.}
% % \footnote{}

\section{Introduction}
\label{sec:intro}

% 动作本身蕴含习惯（examples）
% 然而目前大多数研究专注于精细度和流畅度，很少涉及习惯语义
% image展示，什么是behave-preserved motion transfer
% 
% 主要贡献：
% 1. 考虑到了动作中的习惯，首次提出habit-preserved motion transfer框架,利用category prior flow捕捉物种习惯
% 2. 引入LLM， 实现未知物种的迁移
% 3. 基于deforming things 4D, 提出带骨骼绑定的4D动物数据集
% 

In nature, animals exhibit a remarkable diversity of motion patterns, shaped by their unique anatomy, behavioral traits, and ecological roles. From the stealthy agility of a leopard to the deliberate, grounded movements of an elephant, each species embodies a distinct locomotion style that reflects its behavioral identity. With the growing adoption of virtual environments, video games, and animation-driven storytelling, the demand for realistic animal motion generation and transfer has significantly increased. However, animal motion, especially across different categories, remains relatively underexplored. Manually designing distinct motion styles for every 3D entity is both time-consuming and resource-intensive.

Motion transfer, which adapts a source entity’s motion to a target, offers a promising avenue to streamline this process. Significant strides have been made in human motion synthesis, where two dominant paradigms prevail: motion retargeting and motion style transfer. Motion retargeting focuses on aligning motion data across skeletons with differing topologies, employing constraint-based optimization~\cite{gleicher1998retargetting, lee1999hierarchical, tak2005physically} or deep learning techniques~\cite{aberman2020skeleton, zhang2023skinned, li2022iterative} to ensure structural compatibility. Motion style transfer, conversely, seeks to blend the content of one motion with the stylistic flair of another, often using generative models~\cite{holden2016deep, aberman2020unpaired, song2024arbitrary, kim2024most} to synthesize aesthetically varied outputs. However, these approaches, tailored primarily for human subjects, fall short when applied to animals. They prioritize either structural fidelity or superficial stylistic variations, overlooking the deep-seated physiological constraints and behavioral nuances—termed habitual behaviors—that define animal locomotion.

To address these limitations, we propose habit-preserved motion transfer, a novel paradigm for transferring motion across quadruped species while preserving their physiological and behavioral characteristics, which we term habitual behaviors. Unlike stylistic variations, habitual behaviors reflect rigid physiological constraints and species-specific locomotion strategies. For example, a cat’s joints fold inward when resting, while a cow’s bend outward—a distinction rooted in anatomy rather than style. Unlike motion retargeting, which focuses narrowly on skeletal alignment, our approach ensures that the transferred motion aligns not only at anction-level but also semantically with the target animal’s behavioral identity. Similarly, a cow cannot replicate a cat’s agile gait due to fundamental differences in their movement patterns. This task extends beyond traditional motion retargeting, which aligns skeletal structures, and style transfer, which prioritizes aesthetic variation, by demanding semantic consistency and behavioral fidelity.
These habitual behaviors, encompassing locomotion strategies and joint articulation patterns, are critical to producing authentic motion but are largely ignored by existing methods.

Our approach is grounded in a generative framework based on Vector Quantized Variational Autoencoder (VQ-VAE)~\cite{vqvae}, which incorporates a habit-preservation module alongside category-specific habit encoder. This design enables the model to learn motion priors that reflect the unique behavioral tendencies of different species. In contrast to style, which is often abstract and difficult to interpret, habits are more semantically grounded, offering richer insights into motion patterns and greater potential for downstream applications.
To further enhance the generalization capabilities of our framework, we integrate a large language model (LLM)-based text encoder, allowing the system to incorporate external semantic knowledge. This enables motion transfer even for previously unobserved categories by retrieving habit latents through textual descriptions of category-specific behavior.
Finally, to support this paradigm, we introduce DeformingThings4D-skl, an extended version of the DeformingThings4D dataset~\cite{li2021deformthings4d}, enhanced with detailed skeletal rigging across multiple quadruped species. We conduct extensive motion generation and cross-species transfer experiments on this dataset, demonstrating the effectiveness and semantic fidelity of our method. To the best of our knowledge, this is the first study to explicitly evaluate whether transferred motion is both semantically natural and consistent with the habitual behavior of the target animal.

In summary, our main contributions are as follows:
\begin{itemize}
    \item We introduce the concept of habitual behaviors in motion, and propose a habit-preserved motion transfer framework that incorporates category-specific habit encoders.
    \item We integrate category-specific motion information with text embeddings derived from a LLM to introduce external knowledge, enabling motion transfer to unobserved categories. 
    \item  We present an extended dataset, DeformingThings4D-skl, which includes skeletal bindings for multiple animal species. We conduct comprehensive experiments and quantitative analyses to demonstrate the superiority of our method.
\end{itemize}
\section{Related Works}
\label{sec:related_works}

\subsection{Motion Generation}
Motion generation aims to generate semantically coherent and natural pose sequences~\cite{zhu2023human}. In recent years, most research within this field focuses on generating motions of human~\cite{tevet2022motionclip, dabral2023mofusion, zhang2023t2m-gpt, zhou2023ude, Yang2024omnimotiongpt, lee2019dancing, siyao2022bailando, li2022danceformer, le2023music, yi2023generating, ao2023gesturediffuclip, corona2020context, wang2022humanise, huang2023diffusion} and quadruped~\cite{biggs2019SMAL, Yang2024omnimotiongpt, zhang2024motion} based on conditional signals, such as text~\cite{tevet2022motionclip, dabral2023mofusion, zhang2023t2m-gpt, zhou2023ude, Yang2024omnimotiongpt, zhang2024motion}, audio~\cite{lee2019dancing, siyao2022bailando, li2022danceformer, le2023music, yi2023generating, ao2023gesturediffuclip}, and scene contexts~\cite{corona2020context, wang2022humanise, huang2023diffusion}.
With the advancement of generative models, variational autoencoders (VAE)~\cite{kingma2013auto} and diffusion models~\cite{ddpm} have been widely adopted in motion generation tasks. Zhang et al. ~\cite{zhang2023t2m-gpt} introduced an effective approach for generating motion from text, leveraging a combination of VQ-VAE~\cite{vqvae} and Generative Pre-trained Transformer (GPT)~\cite{gpt, vaswani2017attention} models.
Similarly, Dabral et al.~\cite{dabral2023mofusion} proposed a denoising diffusion-based framework for high-quality conditional human motion synthesis, capable of generating long, temporally plausible, and semantically accurate motions.

These approaches provide a strong foundation for motion generation models. However, while most existing methods focus on achieving high fidelity, diversity, and condition consistency, few have addressed the importance of preserving the behavioral patterns inherent in motion.

% \subsection{Motion retargeting}

% \subsection{Motion style transfer}
\subsection{Motion Transfer}
Motion transfer aims to transfer motion from a source entity to a target entity, playing a pivotal role in animation production, game development, and virtual reality. Building on the foundational concept of direct transfer, researchers have developed more refined approaches, such as motion retargeting and motion style transfer, to handle nuanced sub-tasks.

Motion retargeting~\cite{aberman2020skeleton, zhang2023skinned} focuses on mapping movements across different skeletal templates, allowing for the transfer of motion between characters with varying body shapes and sizes. Aberman et al.~\cite{aberman2020skeleton} introduced a deep learning framework for data-driven motion retargeting between skeletons with distinct structures but topologically equivalent graphs. Zhang et al.~\cite{zhang2023skinned} proposed a residual retargeting network that operates on both skeleton-level and shape-level motion retargeting.
While these methods primarily address the geometric aspects of motion, they do not account for the semantic information embedded in the motion itself.

Motion style transfer~\cite{holden2016deep, aberman2020unpaired, jang2022motion, wen2021autoregressive, song2024arbitrary} blends two motions to create a new motion with one's content and the other's style. Aberman et al.~\cite{aberman2020unpaired} introduced a data-driven framework for unpaired motion style transfer.
Song et al.~\cite{song2024arbitrary} proposed a multi-condition motion latent diffusion model for motion style transfer.
However, they still fail to capture behavioral patterns, which is implicit, shared information derived from multiple samples of a given category.

We propose habit-preserved motion generation and transfer, which focuses on retaining individual or category-specific movement traits during transfer, ensuring that the resulting motion reflects natural behaviors. It emphasizes deeper behavioral understanding, making it more intelligent and adaptable for tasks requiring realistic, habitual motions.

\section{The Proposed Method}
\label{sec:methods}

\begin{figure*}
    \centering
    \includegraphics[width=1\linewidth]{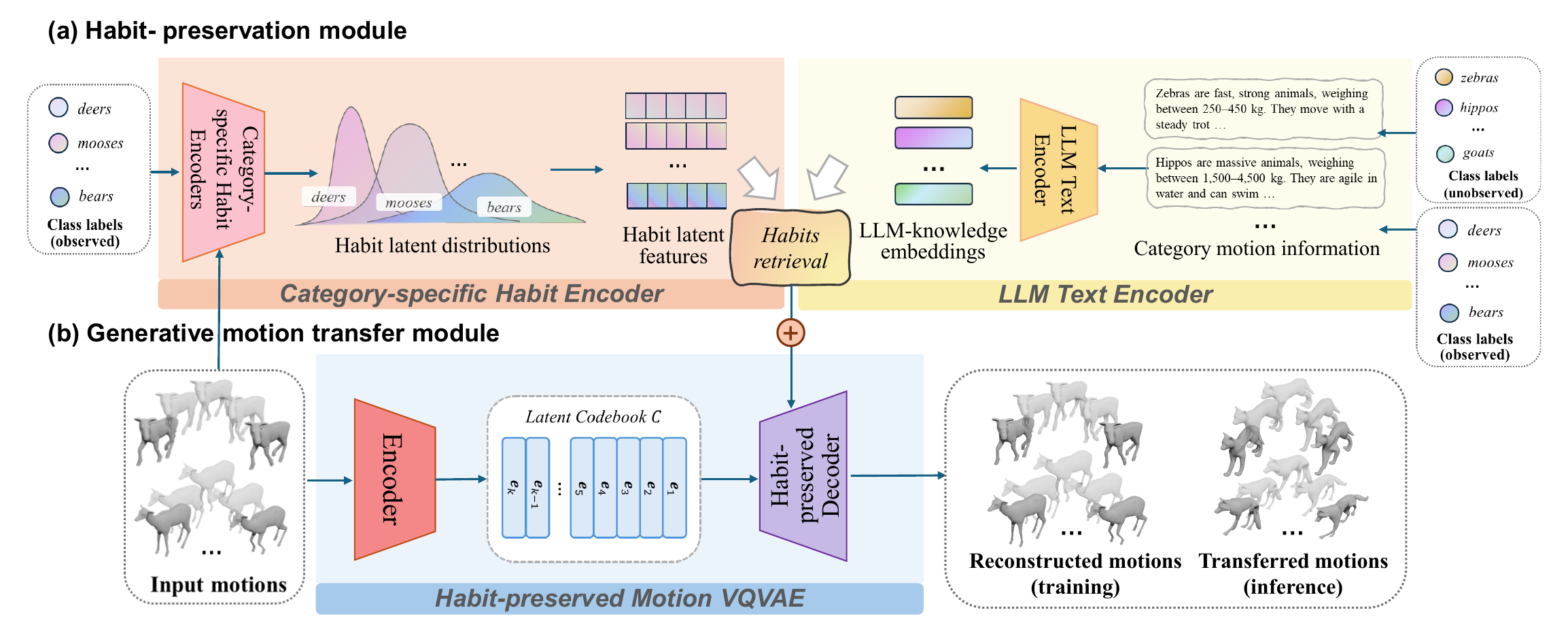}
    \caption{\textbf{The framework of our habit-preserved cross category motion transfer method.} Motion is either reconstructed (during training) or transferred (during inference) through a habit-preserved motion VQ-VAE framework. The decoder takes as input both the latent motion codes and habit features from category-specific habit encoders and LLM text encoders. For observed categories, we employ category-specific habit encoders to extract habit latent features from learned habit distributions. To generalize across categories—including unseen ones—we incorporate an LLM-based text encoder to obtain semantic embeddings that encode category-specific knowledge. By leveraging a habit retrieval mechanism, our framework enables motion transfer even to previously unobserved categories. }
    \label{fig:framework}
\end{figure*}

Our main goal is to transfer motion from a source entity to a target entity while preserving the behavioral habits of the target with high quality. 
As demonstrated in Fig.~\ref{fig:framework}, the overall framework consists of two key modules: the generative motion transfer module and the habit-preservation module. 
In the generative motion transfer module, we use a VQ-VAE as the backbone model, which learns latent motion embedding codes through a discrete process. 
To integrate habit information, we develop a prior flow-based category-specific habit encoder and a LLM text encoder. 
The category-specific habit encoder captures the prior distribution of habits for each category, embedding a habit-specific latent code into each motion. 
The LLM text encoder supplies supplementary motion-related information for both known and novel species, enhancing category-specific representations and enabling motion transfer to previously unseen category.

During training, the model learns by reconstructing the motion of the source entity. 
In the inference phase, it combines the motion of the source entity with the label of the target entity. 
This process incorporates both the motion information and the pre-trained habit prior latent features of the target category, facilitating habit-preserved motion transfer across categories.

We first introduce the problem definition of motion transfer in Sec.~\ref{sec:problem_define}. Next, we present the generative motion transfer module in Section~\ref{sec:motion_transfer}. Finally, in Section ~\ref{sec:habit_preserve}, we introduce the habit-preservation module, which incorporates a category-specific habit encoder and a LLM text encoder.

\subsection{Problem Definition}
\label{sec:problem_define}

For both the source and target motions, each motion $m_t$ for $t=1,2,\dots ,T$ is represented as ${\textbf{m}}_t = \{{\textbf{q}}_t , {\textbf{v}}_t\}$, where $\textbf{q}_t$ and $\textbf{v}_t$ are the local joint rotation quaternions and joint velocities, respectively. Here, $\textbf{q}_t\in R^{N\times 4}$ and $\textbf{v}_t\in R^{N\times 3}$, with $N$ denoting the number of joints. 
For simplicity, we omit the time index $t$.

The local quaternions of the joint rotation $\textbf{q}$ represent the rotations of each joint relative to the local coordinate system of its parent joint.   Joint positions $\textbf{x}\in R^{N\times 3}$ can be obtained from joint rotations $q$ using forward kinematics (FK), and conversely, joint rotations $\textbf{q}$ can be derived from joint positions $\textbf{x}$ using inverse kinematics (IK).
Joint velocities $\textbf{v}$ are calculated as the difference in global joint coordinates $\textbf{x}$ between consecutive frames.

Given a sequential motion data $\textbf{m}_{\text{src}}$ for a source entity belonging to category $c_{\text{src}}$, our objective is to generate a transferred motion $\textbf{m}_{\text{tgt}}$ for the target entity of category $c_{\text{tgt}}$. The generated motion should retain the distinctive motion characteristics associated with category $c_{\text{tgt}}$.

\subsection{Generative Motion Transfer}
\label{sec:motion_transfer}
We employ a generative approach for motion transfer, aiming to preserve the motion characteristics of the source while maintaining the habitual features of the target category. To achieve this, we develop a VQ-VAE-based framework that integrates both the habit latent and knowledge derived from an LLM.

\subsubsection{Motion VQ-VAE}
To achieve controllable and expressive motion synthesis, we adopt a vector-quantized latent space~\cite{vqvae} to model motion priors while incorporating species-specific behavioral patterns through an explicit habit-preserving fusion mechanism. This design decouples structural motion representation learning from habit conditioning, enabling more flexible and modular generation.

Given a motion $\textbf{m}$ from category $c$, along with its habit latent $\textbf{z}_c$ and category information text embedding $\textbf{g}_c$, our goal is to reconstruct the motion sequence via an autoencoder and a learnable codebook $\mathcal{C}=\{\textbf{e}_k\}_{k=1}^K$ containing $K$ codes, where each code $\textbf{e}_k \in R^{d_c}$ and $d_c$ represents the code dimensionality.

We follow the VQ-VAE structure presented in~\cite{zhang2023t2m-gpt}. Let $E$ and $D$ denote the encoder and decoder of the autoencoder, respectively. We calculate the latent feature $\textbf{f}$ of motion $\textbf{m}$ and reconstructed motion $\hat{\textbf{m}}$ as follows:
\begin{equation}
    \textbf{f} = E(\textbf{m}),
    \hat{\textbf{m}} = D(\hat{\textbf{f}}, \textbf{z}_c, \textbf{g}_c),
\end{equation}
where $\textbf{z}_c$ and $\textbf{g}_c$ represent the category-specific habit embedding and LLM text embedding, respectively.

For each latent feature $\textbf{f}_i$, quantization through the codebook $\mathcal{C}$ involves soft sampling rather than selecting the nearest code directly. 

We first calculate the squared distance between $\textbf{f}_i$ and each $\textbf{e}_k$ in codebook $\mathcal{C}$:
\begin{equation} 
\mathcal{D}(\textbf{f}_i, \textbf{e}_k) = \| \textbf{f}_i - \textbf{e}_k \|^2.
\end{equation}

These distances are then converted into probabilities by applying a softmax function with a temperature scaling factor $\tau$:

\begin{equation}
p(\textbf{e}_k \mid \textbf{f}_i) = \frac{\exp\left(-\frac{\mathcal{D}(\textbf{f}_i, \textbf{e}_k)}{\tau}\right)}{\sum_{j} \exp\left(-\frac{\mathcal{D}(\textbf{f}_i, \textbf{e}_j)}{\tau}\right)}.
\end{equation}

Finally, the index $\hat{\textbf{f}_i}$ is sampled from this probability distribution $p(\textbf{e}_k \mid \textbf{f}_i)$.

\subsubsection{Training Objective}
The standard training objective~\cite{van2017neural}, denoted as $\mathcal{L}_{\text{vq}}$,  consists of three components: a reconstruction loss $\mathcal{L}_{\text{re}}$, an embedding loss $\mathcal{L}_{\text{emb}}$ and a commitment loss $\mathcal{L}_{\text{com}}$.
\begin{equation}
    \mathcal{L}_{\text{vq}} = \mathcal{L}_{\text{re}} + \underbrace{\|\text{sg}[\textbf{f}] - \hat{\textbf{f}} \|}_{\mathcal{L}_{\text{emb}}} + \underbrace{\beta \| \textbf{f} - \text{sg}[\hat{\textbf{f}}] \|}_{\mathcal{L}_{\text{com}}}.
\end{equation}
Here, $\beta$ is a hyper-parameter for the commitment loss and $\text{sg}$ is the stop-gradient operator. We set $\beta=0.02$ in our experiments.

Following the loss construction in~\cite{zhang2023t2m-gpt}, we define the reconstruction loss $\mathcal{L}_{\text{re}}$ as a smoothed $\mathcal{L}_1$ loss for both motion reconstruction and velocity regularization, which is defined as:
% \begin{equation}
\begin{align}
    \mathcal{L}_{\text{re}} &= \mathcal{L}_1^{\text{smooth}}(\textbf{m}, \textbf{m}_{\text{re}}) \\
    &= \mathcal{L}_1^{\text{smooth}}(\textbf{q}, \textbf{q}_{\text{re}}) + \alpha \mathcal{L}_1^{\text{smooth}}(\textbf{v}, \textbf{v}_{\text{re}}),
\end{align}
% \end{equation}
where $\alpha $ is a hyperparameter to balance the contributions of position and velocity in the reconstruction loss. We present the ablation study on parameter $\alpha$ in Sec.~\ref{sec:ablation_velo}.

\subsubsection{Architecture and Quantization strategy}
Following Zhang et al.~\cite{zhang2023t2m-gpt}, we adopt a standard CNN-based architecture consisting of 1D convolutions, residual blocks, and ReLU activations. Specifically, we use two residual blocks, with convolutions featuring a stride of 2 for downsampling and nearest-neighbor interpolation for upsampling. The downsampling rate is set to 4, and the dilation rate is configured as 3.

To prevent codebook collapse during training, we use two strategies from prior work~\cite{razavi2019generating}: exponential moving average (EMA) and codebook reset (Code Reset). The EMA is calculated as follows:

\begin{equation}
\mathcal{C}_t \gets \lambda \mathcal{C}_{t-1} + (1-\lambda)\mathcal{C}_t,
\end{equation}
where $\mathcal{C}_t$ is the codebook at iteration $t$ and $\lambda$ is the exponential moving constant.
This smoothing operation reduces fluctuations in training, resulting in more stable and reliable quantization parameters. The codebook reset strategy periodically re-initializes or updates the codebook, helping to avoid poor local minima and allowing the quantizer to adapt to changing data distributions.

\subsection{Habit Preservation}
\label{sec:habit_preserve}
In this section, we present the habit-preservation module, which ensures that the target entity retains its category-specific behavior when transferring motion from the source entity. For each known category, we train a category-specific prior flow model, which generates a habit latent variable to guide motion generation, as detailed in Sec.~\ref{sec:prior flow}.  For unobserved category, we incorporate an LLM text encoder to acquire external knowledge, helping identify the most likely behavioral habits, as described in Sec.~\ref{sec:llm-knowledge}.

\subsubsection{Category-specific habit encoder}
\label{sec:prior flow}
The goal of the category-specific habit encoder is to generate motions that exhibit meaningful category-specific habits, encoded in the latent representation $\textbf{z}$.
We assume that the habit latent $\textbf{z}$ follows a prior distribution, which we parameterize using normalizing flows~\cite{chen2016variational, dinh2016density}.

To capture this distribution, we define an approximate posterior distribution $\textbf{q}_\varphi(\textbf{z}|\textbf{m}_c)$, which serves as the encoder. The encoder maps the input motion  $\textbf{m}_c$  to a distribution over the latent code $\textbf{z}$. We model this posterior as a Gaussian distribution, as is common practice:
\begin{equation}
     \textbf{z} \sim q_\varphi(\textbf{z}|\textbf{m}_c),  q_\varphi(\textbf{z}|\textbf{m}_c) = \mathcal{N}(\textbf{z}| \mu_\varphi(\textbf{m}_c), \sigma_\varphi(\textbf{m}_c) ),
\end{equation}
where $\mu_\varphi(\textbf{m}_c)$ and $\sigma_\varphi(\textbf{m}_c)$ are the mean and standard deviation of the Gaussian distribution, respectively.

The prior distribution $p(\textbf{z})$, governing the latent variable $\textbf{z}$, is parameterized with normalizing flows, specifically using a series of affine coupling layers. These layers act as a trainable bijector $F_\alpha$  mapping an isotropic Gaussian distribution to a more complex distribution. Because this mapping is bijective, the exact probability of the target distribution can be computed via the change-of-variables formula:
\begin{equation}
    p(\textbf{z}) = p_\textbf{w}(\textbf{w})\cdot \| \det \frac{\partial{F_\alpha}}{\partial{\textbf{w}}}\|^{-1},
\end{equation}
where $\textbf{w}=F_\alpha^{-1}(\textbf{z})$. This flexible prior distribution enables the model to capture the complex, species-specific habits in each motion.

For each motion category, we use a Transformer-based architecture for the encoder network $ q_\varphi(\textbf{z}|\textbf{m}_c)$, which is designed to capture the unique motion habits of that category. Training is guided by minimizing the negative log-likelihood function, expressed as:
\begin{equation}
    \mathcal{L}_{\text{prior}}= - \log \left( p_w(w) \cdot \| \det \frac{\partial F_\alpha}{\partial w} \|^{-1} \right).
\end{equation}
% \begin{equation}
%     \mathcal{L}_{\text{prior}} = D_{KL}(q_\varphi(\textbf{z}|\textbf{m}_c) \| p(\textbf{z}) ).
% \end{equation}

This approach uses a separate encoder model for each category, enabling the network to effectively learn category-specific latent representations. By leveraging the category-specific habit latent $\textbf{z}$, we guide the motion transfer process to preserve the unique behavioral traits of each species.

\subsubsection{LLM text encoder} 
\label{sec:llm-knowledge}

\begin{figure}
    \centering
    \includegraphics[width=0.9\linewidth]{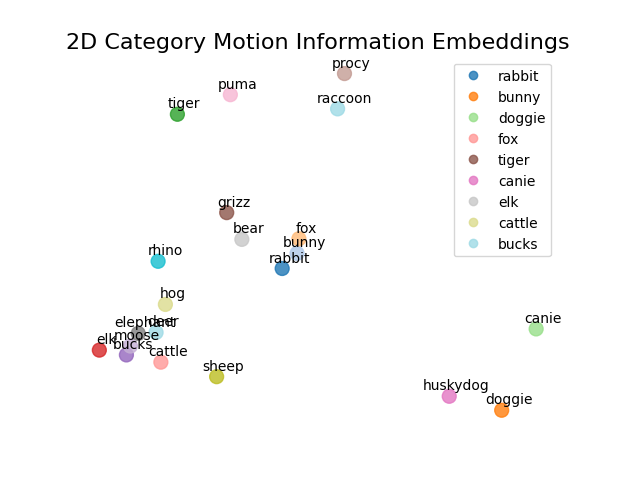}
    \vspace{-0.1cm}
    \caption{\textbf{2D Category Motion Information Embeddings.} Animals with similar body types and behavioral patterns are positioned closer to each other in the diagram.}
    \label{fig:llm_vector}
\end{figure}
\noindent
We integrate an LLM text encoder to acquire external knowledge about the motion characteristics of each category, including those that have not been observed. 
For each category $c$, we gather behavioral information—such as size, weight, speed, limb length, and areas of force exertion using GPT-4o model~\cite{gpt4o}. 
As a knowledge engine, the GPT-4o model provides insights that not only enhance our understanding of category-specific behaviors but also help identify behavioral patterns analogous to those of categories that have not been previously observed. 

To represent the motion information of each category $c$, we utilize BGE embedding~\cite{bge_embedding, cocktail, llm_embedder} to create a text embedding $\textbf{g}_c$. Fig.~\ref{fig:llm_vector} presents a 2D visualization of the text embeddings, illustrating that animals with comparable body sizes and movement patterns are grouped closely together, whereas those with differing characteristics are more distantly located. This clustering offers valuable insights that facilitate the transfer of movement patterns to unseen animals, allowing for the identification of similar behavioral traits.

The text embedding $\textbf{g}_c$ also helps us derive the corresponding habit latent $\textbf{z}_c$. For categories seen during training, $\textbf{z}_c$ can be directly obtained via a pretrained model. However, for unseen categories, i.e., those for which no pretrained model exists to produce $\textbf{z}_c$, we introduce a habit retrieval mechanism to enable generalization.
Specifically, given an unseen category $c$, we leverage its text embedding $\textbf{g}c$ to retrieve the most semantically similar category $c'$ from the set of observed categories $\mathcal{C}{\text{observed}}$, for which pretrained models and corresponding habit latents $\textbf{z}_{c'}$ are available. The similarity is computed in the text embedding space using the Euclidean distance. We then assign the habit latent of $c'$ to the unseen category $c$ as a proxy:

\begin{equation}
\label{eq: unseenz}
    \textbf{z}_c = \textbf{z}_{c'}, c' = \arg \min_{c' \in \mathcal{C}_{\text{observed}}} \, \| \textbf{g}_c - \textbf{g}_{c'} \| _2,
\end{equation}
where $\mathcal{C}_{\text{observed}}$ represents the set of categories for which pretrained models are available.
By grounding the retrieval process in a semantically meaningful space, this approach ensures that the inferred motion behavior for unseen species remains biologically plausible and behaviorally consistent, even in the absence of direct motion data. This makes habit retrieval a crucial component for zero-shot generalization in our framework.

\section{Experiments}
\label{sec:experiments}

\subsection{Experimental Setup}
\subsubsection{Datasets}
\paragraph{DeformingThings4D-skl.}
We present the DeformingThings4D-skl dataset, a new animal motion dataset with skeleton binding and habitual descriptions designed for tasks that require habit-preserved motion transfer. Based on the original DeformingThings4D dataset~\cite{li2021deformthings4d}, DeformingThings4D-skl incorporates skeletal rigging, making it suitable for motion generation, transfer, and analysis tasks. We also provide animal habitual description texts for each category. Using the skeletal structure of a horse with joints number $N=21$ as reference, we applied the automatic rigging tools developed by Baran et al.~\cite{baran2007automatic} and Zhang et al.~\cite{zhang2024magicpose4d} to obtain motion and skin data. To ensure data quality, we selected a subset of DeformingThings4D and performed smoothing and outlier removal on the sequential skeletal data. The final DeformingThings4D-skl dataset consists of 21 distinct animal categories, varying in size and taxonomy, and includes 787 4D motion sequences of totaling 29,505 frames.

\paragraph{AnimalML3D.}
AnimalML3D~\cite{Yang2024omnimotiongpt} is an animal motion dataset that provides multimodal data, including skeletal motion sequences and paired textual motion descriptions. The dataset contains a total of 1,240 animation sequences, each represented with 3D skeletal motion and corresponding textual annotations. We use 922 sequences from 23 animal identities for training and 318 sequences from 13 identities for testing. However, it does not include category-specific behavioral habitual descriptions (for LLM text encoder). To address this limitation, we augment the dataset with additional habit descriptions for each animal category. It is important to note that we only utilize the skeletal motion data in our experiments and do not incorporate the provided motion text annotations.

\subsubsection{Evaluation metrics}
\begin{itemize}
    \item \textbf{Frechet Inception Distance (FID).} We measure the distributional distance between generated and real motion data using the FID score, calculated on motion features extracted by a pre-trained motion Transformer, as proposed in Heusel et al.~\cite{heusel2017gans}.
    
    \item \textbf{Intra-FID.} Proposed by Miyato et al.~\cite{miyato2018cgans}, Intra-FID calculates the FID score between generated and real motions within each specific category, assessing the model's performance at a category level.
    
    \item \textbf{Cross-category Transfer Downstream Task.} To evaluate cross-category motion transfer performance on the DeformingThings4D-skl validation set, we designed a cross-category motion transfer task. This validation set includes motions from 17 distinct animal categories, with each motion transferred to every other category containing at least 5 samples. For each transfer, we calculate the FID score between the transferred motion and the original data of the target category. The average FID score for all transfers serves as a measure of the cross-category transfer capacity of the model.

    \item \textbf{Diversity.} To assess motion diversity within a set, we randomly sample 300 pairs of motions, extract their motion features, and calculate the average Euclidean distance between the pairs. The closer the diversity is to the ground truth, the better.

    \item \textbf{1-Nearest Neighbor Accuracy (1-NNA).} 1-NNA is a metric proposed by Lopez-Paz et al.~\cite{lopez2016revisiting}  to evaluate generative models, assessing whether two distributions are identical. Defined as:
    \begin{equation}
        \text{1-NNA}(S_g, S_r) = \frac{\sum_{X\in S_g}\mathbb{I}[N_X \in S_g] + \sum_{Y\in S_r}\mathbb{I}[N_Y \in S_r]}{|S_g|+|S_r|},
    \end{equation}
    where $\mathbb{I}[\cdot]$ is the indicator function.
    Here, $S_g$ and $S_r$ represent generated and real samples, respectively, and $N_X$ and $N_Y$ are the nearest neighbors of $X$ and $Y$.
    For identical distributions, the accuracy of this classifier should approach 50\% as the sample size increases. An accuracy closer to 50\% implies a higher similarity between $S_g$ and $S_r$, indicating that the model has effectively learned the target distribution.
    \item \textbf{Mean Per Joint Position Error (MPJPE).}  MPJPE computes the average Euclidean distance between predicted and ground-truth joint positions across all frames and joints. In the context of motion transfer, an excessively large MPJPE indicates poor preservation of motion action, while an overly small value may suggest that the transferred motion is overly similar to the original, potentially lacking the intended habitual transformation, especially in cross-category scenarios.

\end{itemize}

\begin{table*}[]
\centering
\caption{\textbf{Comparison results of habit-preserved motion transfer task.}  (gt) means the ground truth results. The $\star$ denotes that the closer the metric is to 0.5, the better.  The $\rightarrow$ denotes that the closer the (gt) metric, the better.}
\vspace{-0.1cm}
\label{tab:compare}
\begin{tabular}{@{}lccccccl}
\toprule
\textbf{Methods}                & \textbf{FID$\downarrow$}         & \textbf{Intra-FID$\downarrow$}   & \textbf{Downstream tasks$\downarrow$} & \textbf{Diversity$\rightarrow$}   & \textbf{1-NNA$\star$}        &\textbf{MPJPE}\\ \midrule
Direct transfer                 &                      /&                      /&                           0.0157&                 6.0876 (gt)     &                      / &/\\
Abermen et al. \cite{aberman2020unpaired} (Style transfer)& 0.1430& 0.2260& 0.2400& 12.8380& 0.8035
&6.0139\\
\textbf{Ours(EMA)}& \multicolumn{1}{c}{0.0532} & \multicolumn{1}{c}{0.1174} & \multicolumn{1}{c}{0.0125}      & \multicolumn{1}{c}{\textbf{5.6994}} & \multicolumn{1}{c}{0.6618} & 1.0098\\
\textbf{Ours(EMA-reset)}&                     \textbf{ 0.0115
}&                      \textbf{0.0568
}&                           \textbf{0.0070}&                      2.1617&                      \textbf{0.6461} &0.3758
\\ \bottomrule
\end{tabular}
\end{table*}

\begin{table*}[]
\centering
\caption{\textbf{Comparison results of habit-preserved motion transfer task on AnimalML3D dataset.} The $\star$ denotes that the closer the metric is to 0.5, the better. (gt) means the ground truth results. The $\rightarrow$ denotes that the closer the (gt) metric, the better.}
\vspace{-0.1cm}
\label{tab:compare2}
\begin{tabular}{@{}lccccccl}
\toprule
\textbf{Methods}                & \textbf{FID$\downarrow$}         & \textbf{Intra-FID$\downarrow$}   & \textbf{Downstream tasks$\downarrow$} & \textbf{Diversity$\rightarrow$}& \textbf{1-NNA$\star$}        &\textbf{MPJPE}\\ \midrule
Direct transfer                 &                      /&                      /&                           0.331&                 2.3790(gt)&                      / &/\\
 Abermen et al. \cite{aberman2020unpaired} (Style transfer)& 0.0372& 0.0492& 0.806& 3.2862& 0.7086&7.5675\\
\textbf{Ours(EMA)}& \multicolumn{1}{c}{0.0102} & \multicolumn{1}{c}{0.0212} & \multicolumn{1}{c}{0.143}      & \multicolumn{1}{c}{1.8466} & \multicolumn{1}{c}{0.5546} & 2.5715\\
\textbf{Ours(EMA-reset)}&                     \textbf{ 0.0097}&                      \textbf{0.0202}&                           \textbf{0.140}&                      \textbf{1.8478}&                      \textbf{0.5498}&2.1645\\ \bottomrule
\end{tabular}
\end{table*}

\begin{table*}[]
\centering
\caption{\textbf{Ablation studies on habit preservation components of our proposed model on DeformingThings4D-skl dataset.}  When both the category-specific encoder and the LLM text encoder are disabled (first row), the framework functions purely as a reconstruction model. As a result, it naturally achieves low FID and Intra-FID scores.   }
\vspace{-0.1cm}
\label{tab:ablation_habit}
\begin{tabular}{ccccccc}
\toprule
\multicolumn{2}{c}{\textbf{Components}}                                                                 & \multicolumn{5}{c}{\textbf{Evaluation metrics}}                                                                                                  \\ \midrule
\multicolumn{1}{c}{\textbf{Category-specific encoder}}                          & \multicolumn{1}{c|}{\textbf{LLM text encoder}}             & \textbf{FID$\downarrow$}                  & \textbf{Intra-FID$\downarrow$}           & \textbf{Downstream tasks$\downarrow$}     & \textbf{Diversity$\rightarrow$}            & \textbf{1-NNA$\star$}                                  \\ \midrule
               \multicolumn{1}{c}{$\times$}                                   & \multicolumn{1}{c|}{$\times$}                          &              0.0009        &          0.0021            &         0.0083             &            0.4174          &        0.9489                                    \\
\multicolumn{1}{c}{\checkmark} & \multicolumn{1}{c|}{$\times$} &{0.0776}                       & 0.1521 &  0.1946 &   4.9825   & 0.6650 \\
                        \multicolumn{1}{c}{$\times$}                      & \multicolumn{1}{c|}{\checkmark} &     0.0391                 &            0.2564          &       0.0111               &    2.1255                  &          0.8718                                  \\
\multicolumn{1}{c}{\checkmark}                    & \multicolumn{1}{c|}{\checkmark} &            0.0115          &           0.0568           &        0.0070            &         2.1617             &              0.6461                            \\ \bottomrule
\end{tabular}
\end{table*}

\subsubsection{Implementation Details}
For the \textbf{VQ-VAE}, the codebook size is set to 512$\times$512. We use AdamW optimizer and a batch size of 32. Training was conducted over 200,000 iterations, with a learning rate schedule of 2e-4 for the first 100,000 iterations and 1e-5 for the final 100,000.  We cropped the motion sequences to $T=10$ for training. To efficiently utilize the time-series data, we applied a 10-fold random sampling with replacement.  The parameters $\alpha$ and $\beta$ in $\mathcal{L}_{\text{re}}$ and $\mathcal{L}_{\text{vq}}$ are set to 0.5 and 0.02, respectively.

For the \textbf{Habit encoder}, we trained a separate transformer-based model for each of the 21 categories. Each transformer contains 4 layers with a hidden dimension of 512, 8 attention heads, and a batch size of 32. 

For the \textbf{LLM text encoder}, we generated motion habit information using the GPT-4o model and embedded the resulting text with the BGE model (version bge-base-en-v1.5). The embeddings were then remapped to a 256-dimensional vector for integration into the model.

\vspace{-0.2cm}
\subsubsection{Compared methods}
To the best of our knowledge, this work is the first to address the task of habit-preserved motion transfer. Since previous approaches primarily focus on human data and did not incorporate species and habit information, direct comparisons with our method are challenging. We evaluate our approach against two baselines: direct motion transfer and style-based motion transfer. For the direct transfer baseline, which applies motion from the source entity directly to the target entity, we adopt the implementation of MagicPose4D~\cite{zhang2024magicpose4d}. For the style-based method, which uses a reference motion as a conditioning signal, we employ the implementation proposed by Aberman et al~\cite{aberman2020unpaired}.

It is important to note that our task differs fundamentally from traditional style transfer. During inference, our method takes the target label as input, without access to target motion data.

% In the \textbf{direct transfer} approach, motion from the source entity is directly applied to the target entity, for which we used the implementation by MagicPose4D~\cite{zhang2024magicpose4d}. Since direct transfer does not involve transformation or generation of motion, we only report results on downstream tasks and diversity metrics. For \textbf{motion style transfer}, we selected the classic approach by Aberman et al.~\cite{aberman2020unpaired}, where the target entity is treated as a motion style to be transferred.
\vspace{-0.1cm}
\subsection{Quantitative Results}
Table~\ref{tab:compare} presents the results of the habit-preserved motion transfer task on the DeformingThings4D-skl dataset, while Table~\ref{tab:compare2} provides the corresponding results on the AnimalML3D dataset. For the direct transfer method, we exclude the FID, Intra-FID, and 1-NNA metrics, as this approach does not employ a generative model. Additionally, in the diversity calculation for direct transfer, diversity is defined as the diversity of the ground truth data.

We compare the performance of the direct transfer and style transfer methods against our proposed approaches. The results labeled "Ours (EMA)" and "Ours (EMA-reset)" reflect two distinct EMA quantization strategies, with the latter incorporating an additional codebook reset mechanism. Our method exhibits substantial improvements over both the direct transfer and style transfer approaches in cross-category motion transfer downstream tasks. Notably, the EMA-reset strategy enhances performance across FID, Intra-FID,  1-NNA, downstream tasks, and MPJPE metrics. Conversely, the EMA strategy yields superior outcomes in diversity metrics.

\subsection{Qualitative Comparison}

\begin{figure*}
    \centering
    \includegraphics[width=0.88\linewidth]{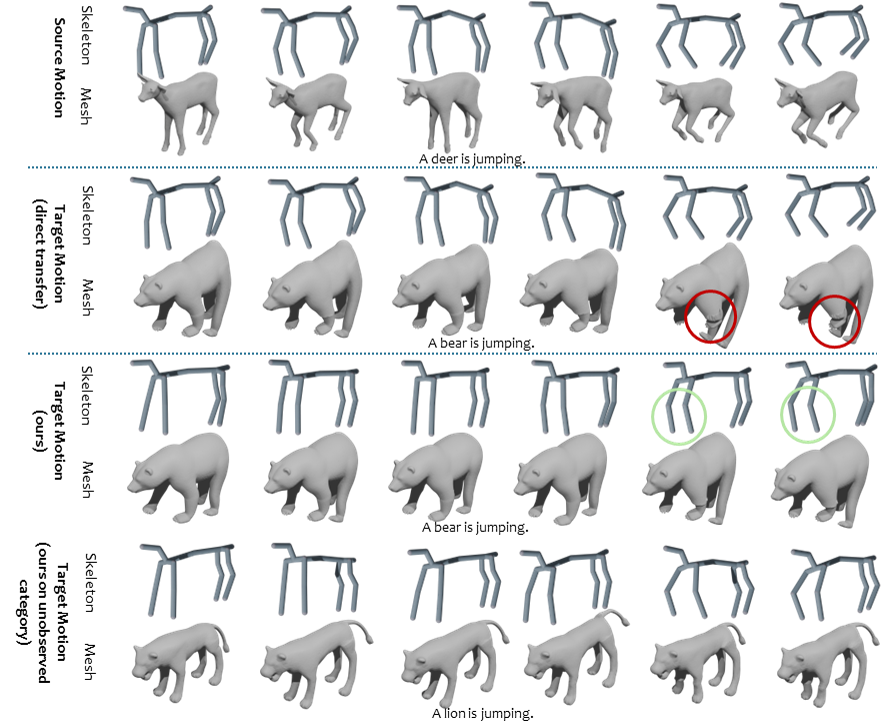}
    \vspace{-0.2cm}
    \caption{Visualization of motion transfer results at the skeleton and mesh levels. The \textcolor{green}{green}  circles highlight how our generated motion adapts to the bear's natural behavioral habits, resulting in less pronounced leg bending. In contrast, the \textcolor{red}{red} circles illustrate unnatural bending angles and model glitches introduced by direct transfer methods. }
    \label{fig:visualization}
\end{figure*}

Fig.~\ref{fig:visualization} shows the visualization of the results of the cross-category motion transfer. We visualize the original 4D data as the source motion, and the result of blending the transferred motion with the target object as the target motion.  For each source motion (first row), we transfer it to both known and unknown species. We compare the results of the direct transfer method on known species (second row) with those of our approach (third row), and we also present the motion transfer results for unseen species (fourth row). As shown, the direct transfer method does not account for species-specific joint angles, muscle habits, and other physiological traits, which can result in unnatural bending in the transferred motion (highlighted in red circles in the figure). In contrast, our method produces motion transfer results that better align with the natural habits of the target species. 
More visualization results are provided in the supplementary materials.

\vspace{-0.2cm} 
\subsection{Ablation Studies}
% ablation on habit components; ablation on feature dim; ablation on velocity feature; ablation on loss \alpha; ablation on soft code

\subsubsection{Ablation on habit preservation}
We conduct ablation studies on the components of habit preservation of our proposed model, the results are shown in Table~\ref{tab:ablation_habit}.
The results show that the model using both the category-specific habit encoder and the LLM text encoder achieves the best performance on downstream tasks. Compared to models that use only the habit encoder or only the LLM text encoder, this combined approach demonstrates improvements across several metrics, including FID, Intra-FID, downstream task performance, and 1-NNA.
When both the category-specific encoder and the LLM text encoder are disabled (first row), the framework functions purely as a reconstruction model. As a result, it naturally achieves low FID and Intra-FID scores. However, it fails to preserve habitual-related motion characteristics, leading to subpar performance in qualitative visualizations, diversity and downstream tasks.
In contrast, our approach not only delivers superior visual quality but also significantly improves diversity.
\vspace{-0.3cm}
\subsubsection{Ablation on velocity loss}
\label{sec:ablation_velo}
\begin{table}[h]
\centering
\caption{Performance metrics for different values of alpha.}
\vspace{-0.2cm}
\begin{tabular}{ccccc}
\toprule
\textbf{alpha} & \textbf{FID} & \textbf{Intra\_FID} & \textbf{downstream task} & \textbf{diversity} \\ \hline
\textbf{0} & 0.046 & 0.219 & 0.078 & 1.893 \\ 
\textbf{0.25} & 0.063 & 0.154 & 0.009 & 2.288 \\ 
\textbf{0.5} & 0.053 & 0.114 & 0.012 & 2.070 \\ 
\textbf{0.75} & 0.037 & 0.156 & 0.017 & 1.904 \\ 
\textbf{1} & 0.037 & 0.163 & 0.092 & 2.007 \\ \bottomrule
\end{tabular}
\label{tab:performance_metrics}
\end{table}
We conducted an ablation study on the velocity loss coefficient $\alpha$ using the EMA quantization strategy to assess the impact of varying this hyperparameter on model performance. Table~\ref{tab:performance_metrics} illustrates how the four key metrics—FID, Intra-FID, downstream task performance, and diversity across different values of $\alpha$ (0, 0.25, 0.5, 0.75, and 1).

As shown, at $\alpha = 0.5$, we observe a favorable trade-off: both downstream performance (0.012) and Intra-FID (0.114) are low, while diversity (2.070) and FID (0.053) are well-balanced. This suggests that moderate velocity regularization encourages both realism and generalization.

% \subsection{Discussion}

\vspace{-0.5\baselineskip}
\section{Discussions}
\label{sec:discussions}
\paragraph{Habit-preserved motion transfer for human.}Habitual behaviors are not exclusive to animals; humans can also exhibit distinct motion habits, depending on factors such as body type and gender. However, existing human motion datasets with clearly distinguishable habitual patterns remain scarce. Most current human motion style transfer datasets rely on exaggerated or explicit motions (e.g., "chicken walk") to represent style. As a result, style transfer tasks in this domain often resemble a blend between two distinct actions, rather than a semantic transformation that captures non-action-related characteristics such as movement rhythm and subtle habitual cues. In contrast, animal motion datasets inherently exhibit greater inter-category diversity and more prominent habitual distinctions, making them a more suitable starting point for exploring habit-preserved motion transfer. For this reason, we first conduct our experiments on animal datasets. Nevertheless, we view extending this line of research to human motion data as a valuable direction for future work.

% \paragraph{}
% \vspace{-0.4cm}
\section{Conclusion}
\label{sec:conclusion}
% we tackle the challenge of preserving category-specific habits in cross-category motion transfer. We propose a habit-preserved method for motion generation and transfer, building on the generative VQ-VAE framework.  To preserve category-specific traits, we introduce a category-specific habit encoder pretrained with prior flow. Additionally, we incorporate an LLM-based text encoder to provide external information, enabling motion transfer to unseen categories. We also introduce the DeformingThings4D-skl dataset and conduct cross-category motion transfer experiments on this dataset. The experimental results demonstrate the superiority of our method.
% In the future, we aim to extend our model's capability to handle motions with different skeleton architectures, enhancing its universality in motion transfer.

We propose a habit-preserved motion generation and transfer framework based on VQ-VAE, integrating a habit encoder pretrained with prior flow to preserve category-specific traits, and an LLM-based text encoder to provide external information, thus supporting unseen category motion transfer. We introduce the DeformingThings4D-skl dataset and validate the effectiveness through comprehensive experiments. Future work will explore generalization to motions with diverse skeleton architectures.
\clearpage
{
    \small
    \bibliographystyle{ACM-Reference-Format}
    \bibliography{sample-sigconf-authordraft}
}

% \appendix
% % \section{Research Methods}

\end{document}